\documentclass{article}

    \usepackage[preprint]{neurips_2020}
  \usepackage{algorithm}
\usepackage{dsfont,microtype}

\usepackage{algorithm,algorithmic}

\usepackage{graphicx}
\usepackage{tabularx}
\usepackage{color}
\usepackage{tikz}
\usepackage{subfigure}

\usepackage[utf8]{inputenc} 
\usepackage[T1]{fontenc}    
\usepackage{hyperref}       
\usepackage{url}          

\usepackage{pgf}
\usepackage{bbm}

\usepackage{graphicx}
\usepackage{placeins}
\usepackage{booktabs}       
\usepackage{amsfonts}       
\usepackage{nicefrac}      
\usepackage{algorithmic}
\usepackage{microtype}      
\usepackage{amsmath}
\setcitestyle{numbers}
\setcitestyle{square}

\DeclareMathOperator*{\argmax}{arg\,max}

\title{Statistical Consequences of Dueling Bandits}

\author{%
  Nayan Saxena \\
  Department of Statistical Sciences\\
  University of Toronto \\
  \texttt{nayan.saxena@mail.utoronto.ca} \\
  \And
  Pan Chen\\
  Department of Computer Science\\
  University of Toronto\\
  \texttt{pan.chen@mail.utoronto.ca} \\
  \And
 Emmy Liu \\
Department of Computer Science,\\ Cognitive Science Program\\
  University of Toronto\\
  \texttt{me.liu@utoronto.ca} \\

}

\begin{document}

\maketitle

\begin{abstract}
Multi-Armed-Bandit frameworks have often been used by researchers to assess educational interventions, however, recent work has shown that it is more beneficial for a student to provide qualitative feedback through preference elicitation between different alternatives, making a dueling bandits framework more appropriate. In this paper, we explore the statistical quality of data under this framework by comparing traditional uniform sampling to a dueling bandit algorithm and find that dueling bandit algorithms perform well at cumulative regret minimisation, but lead to inflated Type-I error rates and reduced power under certain circumstances. Through these results we provide insight into the challenges and opportunities in using dueling bandit algorithms to run adaptive experiments.
\end{abstract}


\section{Introduction}

Rapid growth in online learning has provided scientists and researchers a new digital platform to conduct randomised experiments with students in real-world settings. This data can be leveraged by education researchers to adaptively assign students carefully tailored educational material,  and further improve the quality of their content through careful exploratory assessment of the assigned conditions \cite{rafferty2019statistical, williams2018enhancing}. To achieve these goals, one of the most popular problem frameworks is the Multi-Armed Bandit (MAB) problem which focuses on optimal assignment of conditions based on numerical reward signals from the participant (student) whose goal is to trade-off between exploration and exploitation of conditions \cite{auer2002using,robbins1952some}.  Historically, MAB algorithms have been used in industrial settings to leverage user feedback and adaptively display more popular advertisements, websites and produce content \cite{chakrabarti2008mortal, kohli2013fast}, whereas in research settings these algorithms have been used for clinical trials \cite{kuleshov2014algorithms, sui2014clinical,bather1980randomised}, robot control \cite{matikainen2013multi, mcguire2018failure}, and by behavioral and social scientists for crowd-sourcing experiments \cite{abraham2013adaptive, berkowitz2018one, athey2017efficient}. 

While MAB algorithms focus entirely on quantitative feedback (scalar reward), it has been shown that human motor learning is maximised when subjected to both qualitative and quantitative feedback \cite{kilduski2003qualitative} and recent work suggests that quantitative metrics might not be the best indicators of true human preferences \cite{ radlinski2008does}. Therefore, a more qualitative approach for preference elicitation from human participants should be more ideal to leverage student feedback, which can in turn be achieved through preference based MAB (dueling bandit) algorithms like Double Thompson Sampling (DTS) \cite{busa2014survey,wu2016double}, which focuses on participants simply choosing between two presented alternatives.

Through this paper, we take a closer look at leveraging these dueling bandit algorithms for adaptive experimentation by empirically investigating the statistical properties of the DTS algorithm against uniform random experimental condition selection. Our study is primarily motivated by prior work assessing the statistical properties of MAB algorithms in an educational setting \cite{rafferty2019statistical},  and builds upon recent work exploring the challenges that come with conducting hypothesis tests to analyze data from adaptive experiments using bandit algorithms \cite{villar2015multi}, such as proposing strategies for modifying MAB algorithms to trade-off reward and power \cite{yao2020power} and improving coverage of confidence intervals \cite{deshpande2018accurate}.  

In summary, our main contributions are:
\begin{enumerate}
    \item We emphasise the conceptual significance of quantifying the differences in statistical power, regret and false positive rate between uniform sampling and Double Thompson Sampling
    \item Through simulation experiments we show that for a reasonable decrease in power, it may be advantageous to run experiments with dueling bandits when the number of arms is small and the expected effect size is large
    \item By applying the same analysis on the real-world Microsoft Learning to Rank (LTR) dataset we confirm that using DTS results in lowered power, regret and a higher proportion of people being assigned to the better alternative.
\end{enumerate}

\section{Preliminaries}


\subsection{Multi-armed Bandits}
As a precursor to dueling bandits, we consider a central problem in sequential decision making and adaptive experiment design known as the multi-armed bandit (MAB) problem 
\cite{robbins1952some}. The problem setup consists of $k$ arms (actions) such that for every action $a_i \in \mathcal{A}$  the corresponding probability of yielding a success is $p_i\in [0,1]$. With no a priori information about these success probabilities, the central goal for an agent is to maximise the overall number of successes by pulling these arms, or performing these actions, and ultimately settling on one over a fixed period of time denoted by $t$. Assume that under this framework the expected reward of arm $a_i$ is given by $\mu(a_i)$ and $\mu^* = \argmax \mu(a_i)$, for $i\in \{1\ldots k\}$. The objective of an agent is to then minimise the cumulative regret,
$$
    \mathcal{R}^{\text{\tiny MAB}}_t = \sum_{i=1}^{t} [\mu^* - \mu_{a_{(i)}}] 
$$

Here the agent is allowed to only choose one action, $a_{(i)}$ at a given time step. However, in many cases during adaptive experimentation, it may be difficult to frame the result of an action as a success or failure. In particular, in many cases what experimenters are interested in is simply which arm is preferred to the others, in which case pairwise preferences may be more appropriate for the subject to make. To accommodate pairwise preferences we now consider a generalised form of the MAB problem also known as the dueling bandit problem. 

\subsection{Dueling Bandits} 
The dueling bandit problem can be characterised as a special case of the popular Multi-armed Bandit (MAB) problem that focuses on 
pairwise comparisons between actions at every iteration \cite{yue2012k}. The problem setup is similar to the MAB problem, except at every iteration the agent chooses two actions $a_m, a_n \in \mathcal{A}$ and performs a comparison before choosing one action that they prefer \footnote{The traditional dueling bandit framework allows comparisons between  the same actions (self-dueling) where $m=n$, but throughout this paper we assume $m \neq n$  as it is unlikely for users to be presented with comparisons between the same actions during adaptive experimentation .}.   Throughout this paper we specifically utilise the Double Thompson Sampling (DTS) algorithm, which is an adaptation of the popular Thompson Sampling algorithm in the regular MAB context \cite{wu2016double}. A comprehensive outline of the DTS procedure can be found in Appendix A.

Formally, the probability of one arm winning over another is given by ${P}(a_m \succ a_n)$ which we abbreviate as $P_{mn} \equiv {P}(a_m \succ a_n)$.  For each trial $i$ the outcome of each comparison is binary $x_i\sim \text{Bernoulli}(P_{mn})$ where the probability of one arm winning is formally given by,
$$
    {P}(a_m \succ a_n) = \Delta(a_m, a_n)+ 0.5 
$$
Here, $\Delta(a_m, a_n) \in [-0.5,0.5]$ is the difference measure between the two actions, abbreviated as $\Delta_{mn} \equiv \Delta(a_m, a_n) $, such that $a_m \succ a_n \Longleftrightarrow \Delta_{mn}>0$. It is further assumed that the probabilities representing user preferences are stationary over time and modelled by the unknown preference matrix denoted by $\mathbf{P} = [P_{mn}]_{k\times k}$
where the entries satisfy $P_{mn}+P_{nm} = 1$.  

The goal of an agent operating in this environment is to choose the optimal \textit{winning} arm when conducting pairwise comparisons and ultimately minimise regret.  Motivated by classical voting theory, there are different notions of a \textit{winning} arm like the Borda winner \cite{jamieson2015sparse} and Copeland winner \cite{zoghi2015copeland,copeland1951reasonable}.  In this paper we focus on using the notion of a  Condorcet winner \cite{ black1958theory} and strong regret \cite{wu2016double} as described below.


\paragraph{Condorcet Winner}
The Condorcet winner is a single action $a^*$ that beats all other actions in a pairwise comparison, such that $\Delta_{a^{*}a} > 0$ for all $a \neq a^*$. The Copeland winner, found by maximising the normalised Copeland score $ \frac{1}{k-1}\sum_{m\neq n} \mathbbm{1}[P_{mn}>0.5]$ is a Condorcet criterion, which means that it always finds the Condorcet winner if one exists. It should be noted that it is possible for no Condorcet winner to exist, in which case the same definition can be extended to obtain a set of Copeland winners which always exist.
\cite{zoghi2015copeland, wu2016double}.

\paragraph{Strong Regret} Suppose $a^{(i)}_m$ and $a^{(i)}_n$ are actions chosen at a given timestep $i$ and $a_o$ is the optimal action. Then under this framework the overall objective for an agent is to minimise the cumulative strong regret,
$$
      \mathcal{R}^{\text{\tiny DUEL}}_t = \sum_{i=1}^{t} [\Delta_{om}+\Delta_{on}] 
$$

\section{Methods}

\subsection{Experimental Setup}
 In order to examine the impact of dueling bandit algorithms as compared to uniform random assignment during adaptive experimentation, we looked at several conditions relevant to experimenters by varying the number of arms and the effect sizes between pairs of arms.  For each set of conditions 5000 simulations were carried out, each corresponding to a real-world study, where data at each timestep would correspond to  data collected from a participant exposed to a treatment arm. Furthermore, to assess the statistical quality of each algorithm, we considered the statistical power in comparing pairs of arms, the regret accumulated, and the false positive rate. 
 During its operation, each simulation assigned hypothetical participants using either the DTS algorithm, or uniformly at random and was initialized using a preference matrix $P \in \mathbbm{R}^{n \times n}$, where $n$ is the number of arms. For each given pair of arms $(i, j)$, the effect size was calculated (Cohen's w of 0.1, 0.3 or 0.5) of the comparison based on the difference (0.05, 0.15, or 0.25) between $P_{ij}$ and $P_{ji}$, noting that $P_{ij} + P_{ji} = 1$. Finally, in order to compute the false positive rate, preference matrices with zero effect size for all comparisons between pairs of arms were also used.  These experimental conditions are also summarized in Table 1.
 Data and code are publicly available at \url{https://bit.ly/35J2xH0}.


\subsection{Experimental Dataset}
In addition to simulated data, we also present experiments using data collected from users in a real-world context. In particular, we perform simulations using the Microsoft Learning to Rank (LTR) dataset which presents pairs of search queries and documents, along with the relevance ratings of the document~\cite{ltr}. As A/B testing is commonly used to test ranking and recommendation algorithms, this is an appropriate context in which we benchmark outcomes of such experiments. We used an implicit preference matrix over all 136 rankers derived from this dataset similar to~\cite{zoghi2015copeland}, and then randomly sampled Condorcet and non-Condorcet submatrices from this larger matrix in order to simulate experiments comparing different rankers.
The sample size is set as ${n \choose 2} \times m$, where arm $i$ and arm $j$ are the pair of arms with the smallest effect size, and $m$ is the number of participants needed to achieve expected statistical power of 0.8 for effect size 0.1. This simulates the real-world scenario where the number of required participants is unknown. To further evaluate the long-term performance of the two assignment methods, we ran the simulations with sample sizes up to 10 times more than the initial sample size and tracked the same set of metrics as in the main analysis.
\begin{table*}[!htbp]
{
  \centering
  \resizebox{0.8\columnwidth}{!}{
  \begin{tabular}{ll | l}
    \toprule
    \textbf{Condition} & & \textbf{Description} \\
    \midrule
    Sampling type     &      &  Double Thompson Sampling \\
     & & Uniform Random Sampling \\
    \midrule
    Number of arms & & $n \in \{3, 5\}$ \\
    \midrule
    Sample sizes &  Uniform effect sizes & Non-zero effect sizes: $s = 10mn$ where $m$ is the number of participants \\ 
    & & needed to achieve expected statistical power of 0.8 and $n$ is \\
    & & the number of paired comparisons in that simulation     \\
    \cmidrule(r){2-3}
     &  Zero effect size & Same sample sizes as in the non-zero effect case calculated using the\\ 
     & & number of participants needed to achieve expected statistical power of\\ 
     & & 0.8 for effect size 0.3.\\
    \cmidrule(r){2-3}
     &  Learning to Rank dataset & Different effect sizes: $s = 10mn$, where $m$ is the number of participants\\ & & needed to achieve expected statistical power of 0.8 for 0.1 effect size,\\ & & and n is the number of paired comparisons in that simulation.\\

    \midrule
    Effect sizes     & Simulated dataset &  None (0), Small (0.1), Medium (0.3), Large (0.5). Same effect size between\\
    & & each pair of arms. The winning arm in each pair is randomly assigned\\
    & & in non-zero cases. \\
    \cmidrule(r){2-3}
         & Learning to Rank dataset       &   For each pair of arms $i$ and $j$, the effect size is in $[0, 1)$.\\
    \bottomrule
  \end{tabular}}
  \label{tab:table}
   \caption{Summary of conditions varied across simulations}}

\end{table*}

\subsection{Analysis}
After each simulation, we shift our focus to analyzing several outcomes relevant to experimenters: (a) statistical power to detect an effect between each pair of arms, (b) false positive rate in detecting effects when none exist, (c) regret over time, and (d) percentage of participants assigned to the Condorcet winner when one exists.
For simulations where there was some difference between pairs of arms, a Chi-squared contingency test was performed for each pair of arms, with significance level  $\alpha = 0.05$, which is the standard across multiple domains including educational experiments. For each simulation, regret over time, reward over time, and percentage of participants assigned to the Condorcet winner was recorded and aggregated. Finally, to calculate the false positive rate, we considered a pair of arms as a false positive if the comparison between them reached significance, given there should be no effect between these two arms.

\section{Results}

\subsection{Synthetic Data}


\subsubsection{Conditions that differ in terms of effectiveness}

\paragraph{Average Power Over Time}

 In educational settings, different conditions, such as sample solutions, may have a different impact on students' learning efficiency and engagement with the educational resources. However, one condition that outperforms the others is often not statistically significant until testing is conducted with many students involved. Yet, different ways of conducting the experiments may require a different number of participants to reach significant effects amongst conditions.

 \begin{figure}[!htbp]
\centering

\mbox{ \includegraphics[width=0.8\columnwidth]{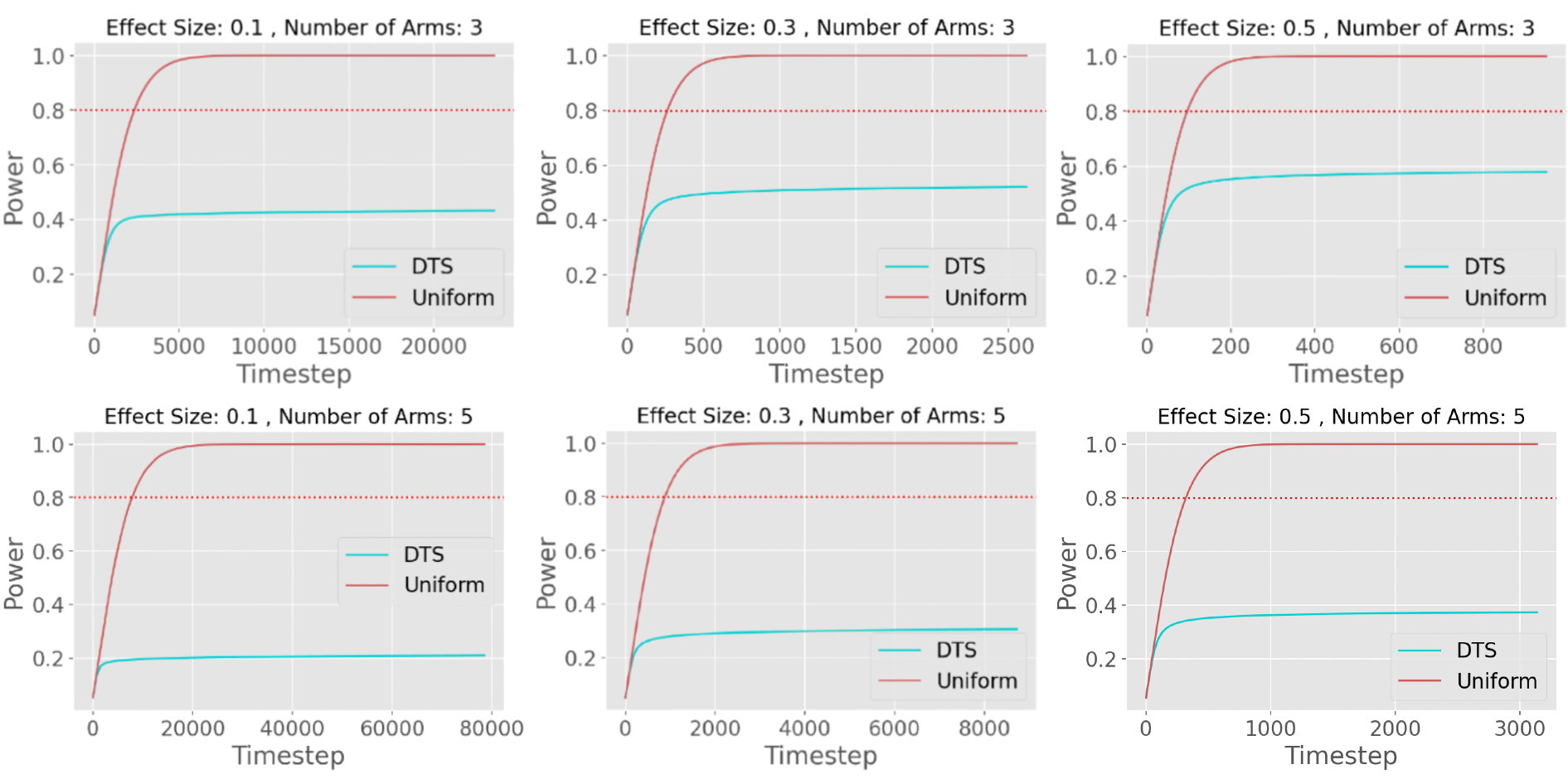} \hskip 3ex}
\caption{Average power over time for DTS algorithm (blue) and Uniform assignment (red) with synthetic data.}
\end{figure} 
 

We modeled the means of the power of different pairs of conditions, as this measures how well each assignment did in finding statistically significant effects among all the conditions. It can be observed in Figure 1 that uniform sampling consistently reached the 0.8 power threshold with fewer participants as compared to DTS, in both 3 and 5-condition settings. This pattern was also observed when final power was recorded for both uniform sampling and DTS, as seen in Figure 2, thus indicating that DTS is more susceptible to inflated Type-II error rates (reduced power). This shows that when there are insufficient students to test among multiple variants, uniform sampling is more often a better way of assignment given the effects among different conditions are of the highest priority.

\begin{figure}[h!]
\centering
\mbox{\hskip -4.3ex \includegraphics[width=0.82\columnwidth]{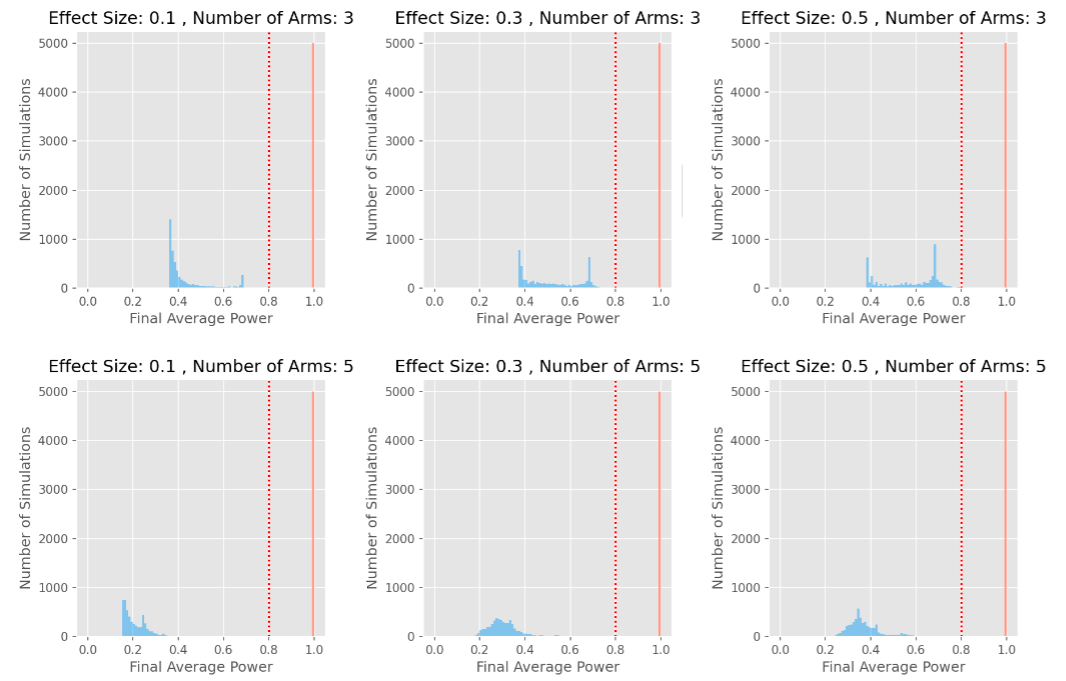} }
\caption{Comparison of final power  between DTS algorithm (blue) and Uniform assignment (red) for simulated dataset where the dotted line indicates effect detected.}
\end{figure}

\paragraph{Proportion of Condorcet Winners} In educational experiments, given that there is one unknown best condition, we examined how uniform sampling and double Thompson sampling fulfill the goal of providing students with the best condition by calculating the proportion of trials in which a student would be assigned to the Condorcet winner.

 \begin{figure}[h!]
\centering

\mbox{\hskip -2.5ex \includegraphics[width=0.8\columnwidth]{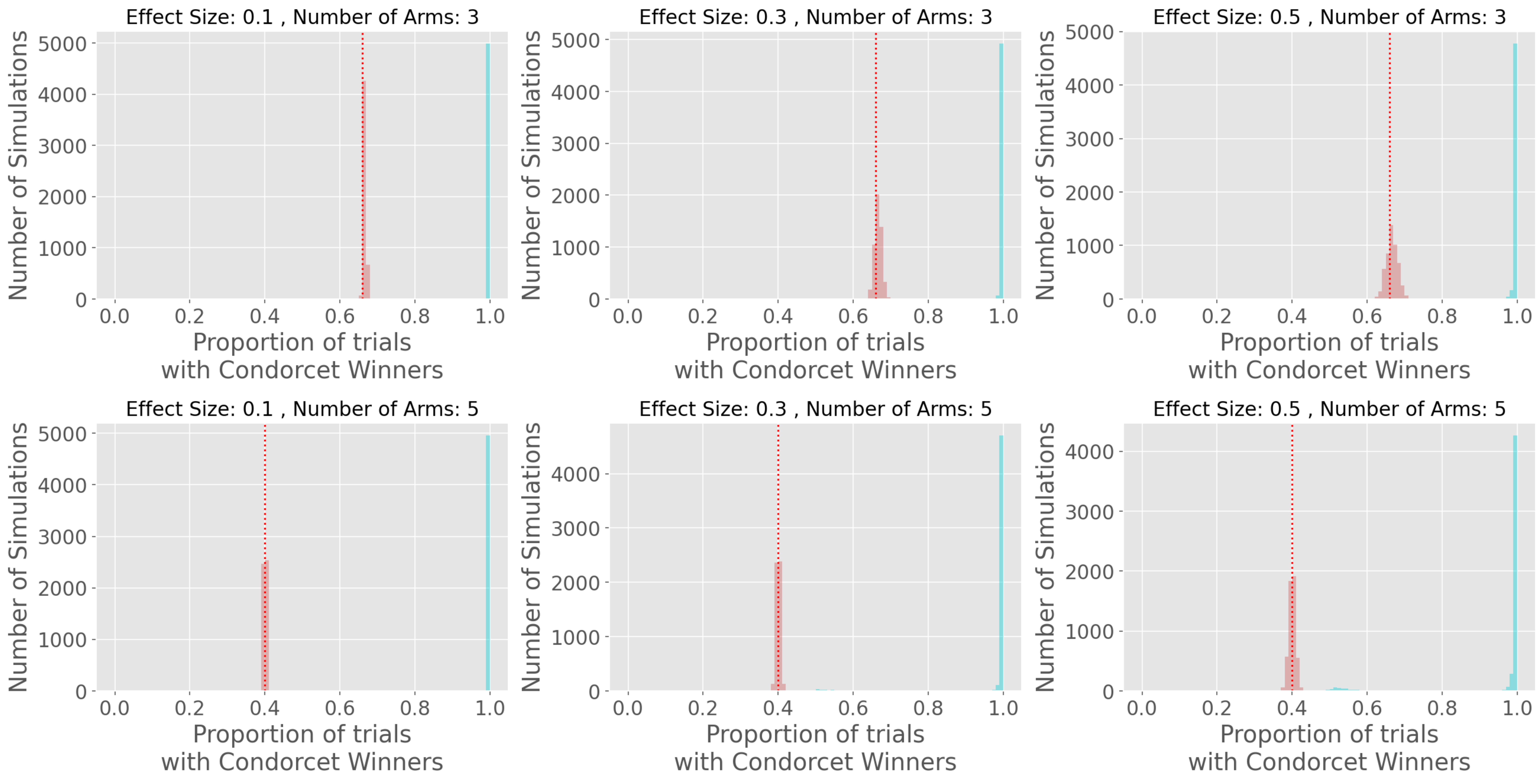} }
\caption{Proportion of trials with Condorcet winners for DTS algorithm (blue) and Uniform assignment (red) with synthetic data. Here, the dotted line indicates expected assignment.}
\end{figure} 

From Figure 3, it can be observed that, on average, DTS assigned a higher number of Condorcet winners to students who participated in the experiments. 
Across both 3 and 5-condition simulations, we observe that DTS assigned the Condorcet winner to almost every student, whereas with uniform sampling the number of students receiving the Condorcet winner is in line with the statistical expectation.


\paragraph{Cumulative Strong Regret} Apart from the proportion of Condorcet winners, we used cumulative strong regret to measure the different experiences uniform sampling and DTS give to students. As seen in Figure 4, a noticeable difference in cumulative strong regret between uniform sampling and DTS  can be seen, which grew more evident with increase in effect size from 0.5 to 0.1, and number of arms from 3 to 5. Across all simulations, DTS performed better by accumulating less regret as compared to uniform sampling.

\begin{figure}[h!]
\centering
\mbox{\hskip -4.3ex \includegraphics[width=0.8\columnwidth]{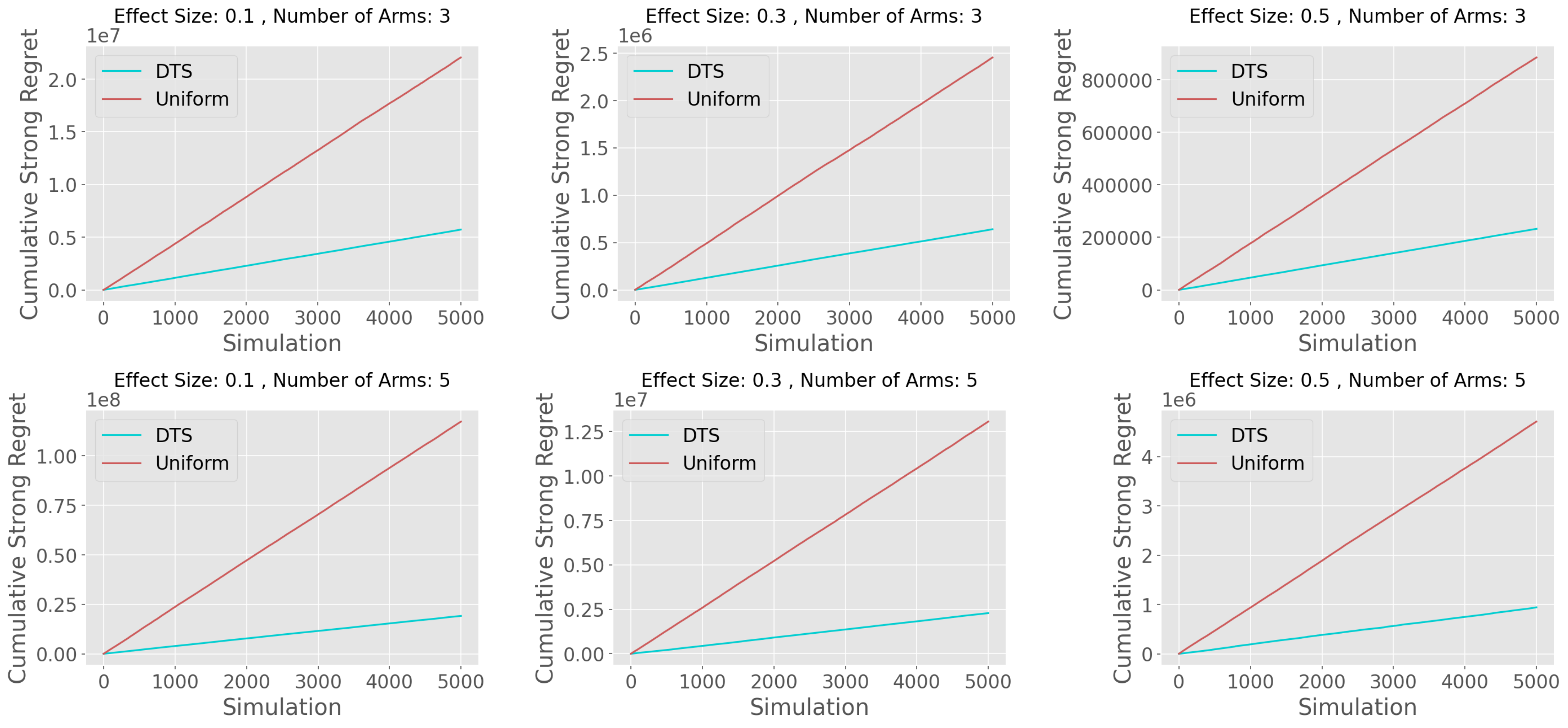} }
\caption{Comparison of cumulative strong regret over time between DTS algorithm (blue) and Uniform assignment (red) for simulated dataset.}
\end{figure} 


\subsubsection{Conditions that are equally-effective}



\paragraph{False Positive Rate}
To simulate scientific settings like clinical trials, along with behavioral and social sciences where little to no difference between arms is commonly observed, we measured the false positive rate when conditions are equally effective. False positives here mean that at least one comparison between arms produces a significance value $< 0.05$ while the effect size between each pair of arms is 0. From Figure 5, we observe that the false positive rates for uniform sampling remain consistent across 3-condition and 5-condition scenarios and are relatively lower when compared to DTS algorithm.


 \begin{figure}[h!]
\centering

\mbox{\hskip -4.2ex \includegraphics[width=0.4\columnwidth]{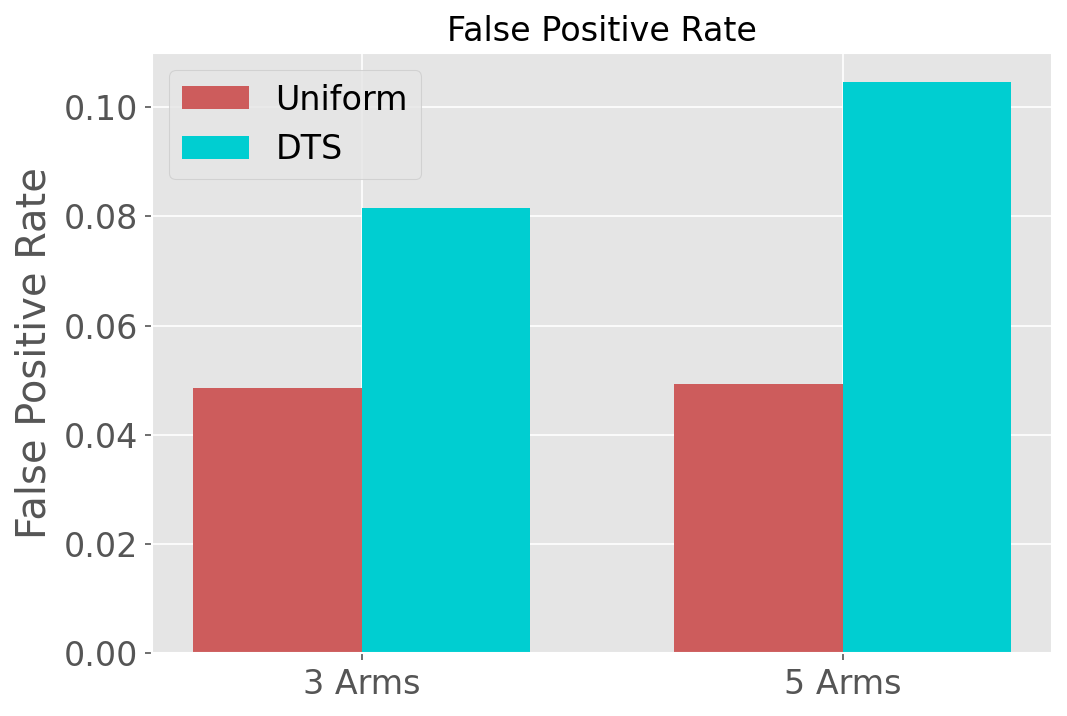} }
\caption{False positive rate for DTS algorithm (blue) and Uniform assignment (red) with simulated data.}
\end{figure} 

This indicates that when there is little to no difference between conditions the DTS algorithm will result in a higher Type-I error as compared to uniform sampling. 
These issues are especially problematic for scientific research, since erroneously believing an intervention is better than a control may result in lack of reproducibility. 


\subsection{Learning to Rank Dataset}

In real settings the effect sizes between any two conditions are not limited to the set $\{0.1, 0.3, 0.5\}$, therefore we ran the same experiments on the LTR dataset. Overall, our results on the real-world dataset were consistent with our simulation studies. We noticed that DTS provided the best condition to students more often and accrued relatively lower average power as seen in Figure 6. It should further be noted that in a few trials DTS did not assign the Condorcet winner to the majority of students possibly because the differences between the best condition and other arms is miniscule, thus requiring more participants for the algorithm to learn the best condition.


\begin{figure}[htp]
\centering

\subfigure{\hskip -3.5ex\includegraphics[width=0.5\textwidth]{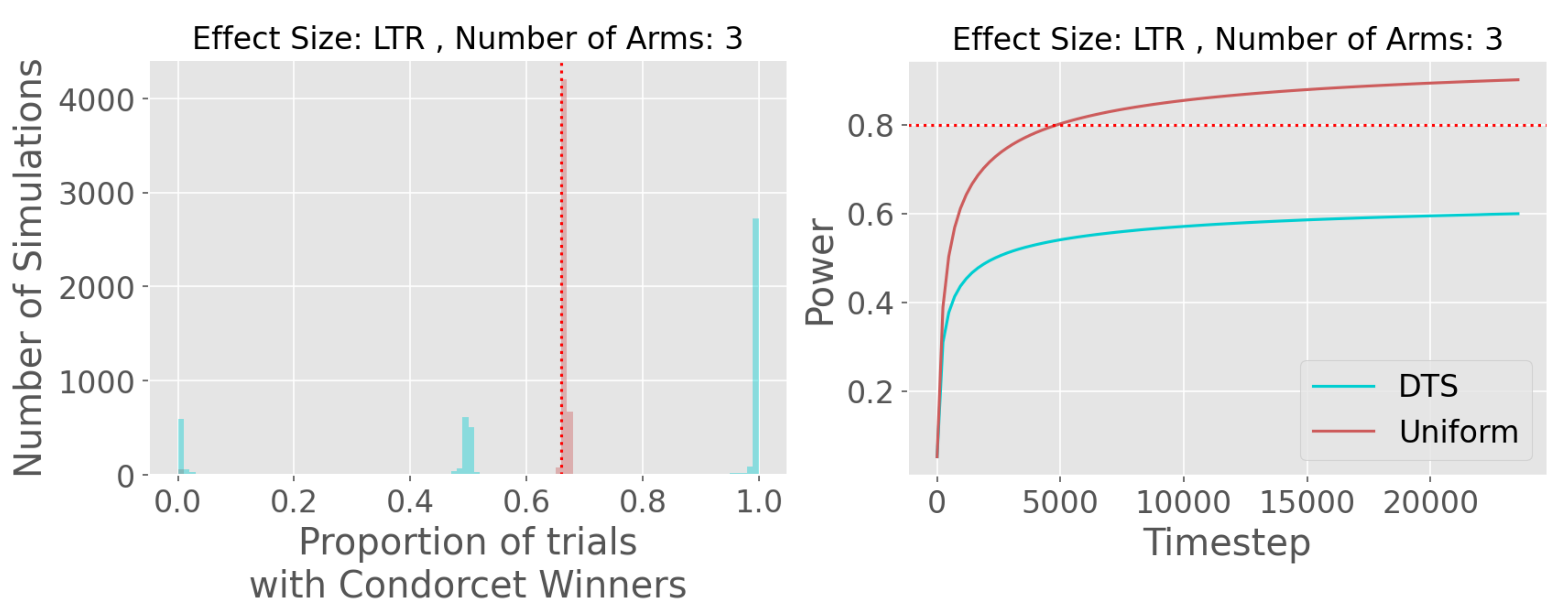}
}

\subfigure{\hskip -3.5ex\includegraphics[width=0.5\textwidth]{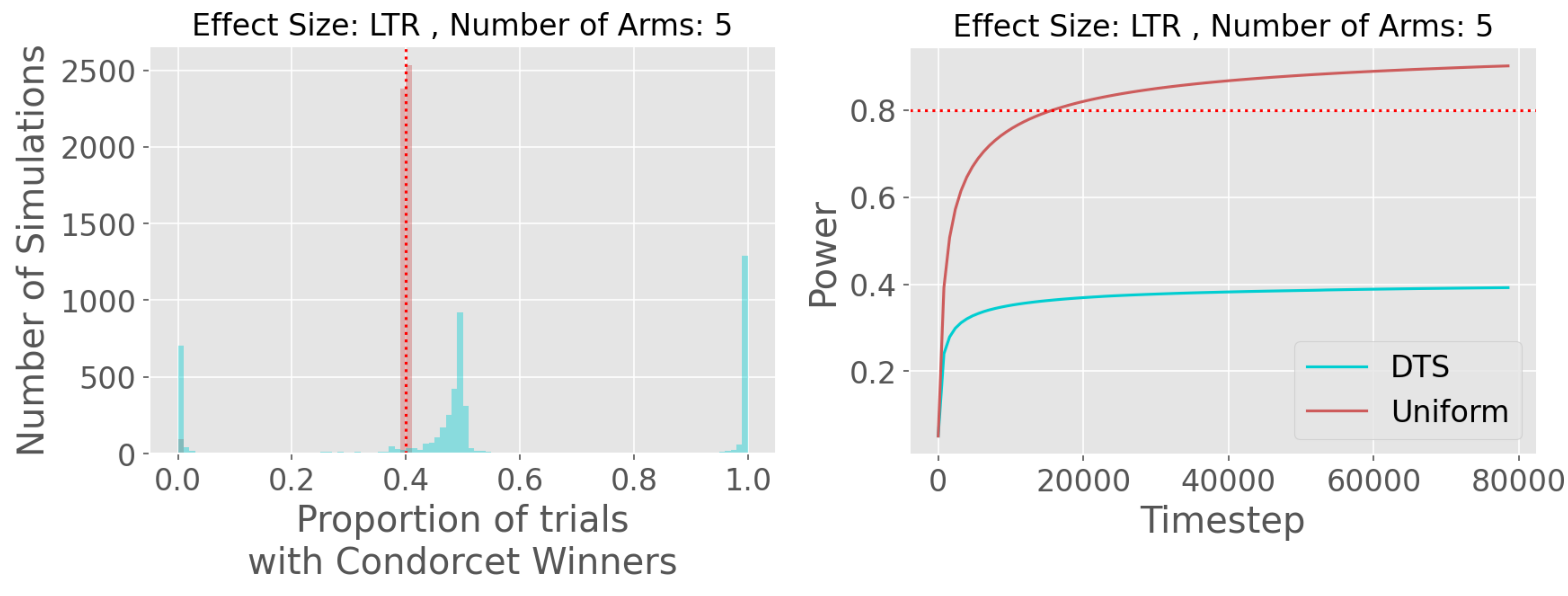}}
\caption{Proportion of trials with Condorcet winners alongside average power over time for DTS algorithm (blue) and Uniform assignment (red) with LTR dataset. } \label{fig2}
\end{figure}


 Consistent with our results on synthetic data, it can further be observed in Figure 6 that uniform sampling reached the 0.8 power threshold with fewer participants as compared to DTS across both 3 and 5-condition settings. This pattern was also observed when final power was recorded for both uniform sampling and DTS, as seen in Figure 7, thus reinforcing the finding that DTS is more susceptible to reduced power even in real world settings.

 \begin{figure}[htp]
\centering
\hskip -1ex
\mbox{\subfigure{\includegraphics[width=0.32\textwidth]{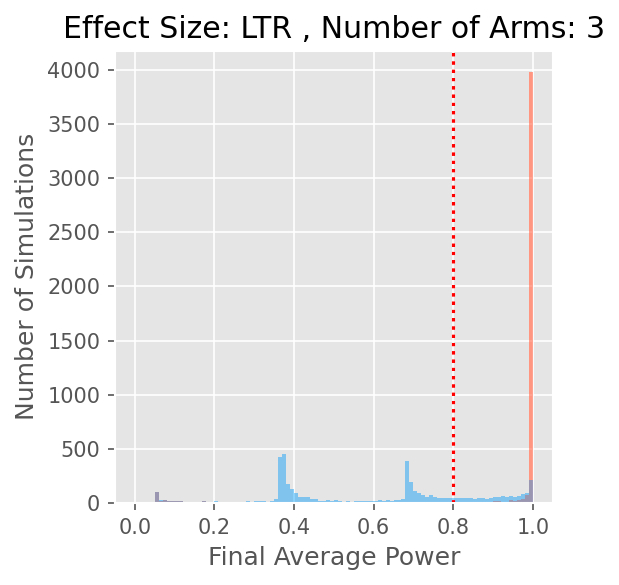}} 
\hskip -0.2ex
\subfigure{\includegraphics[width=0.32\textwidth]{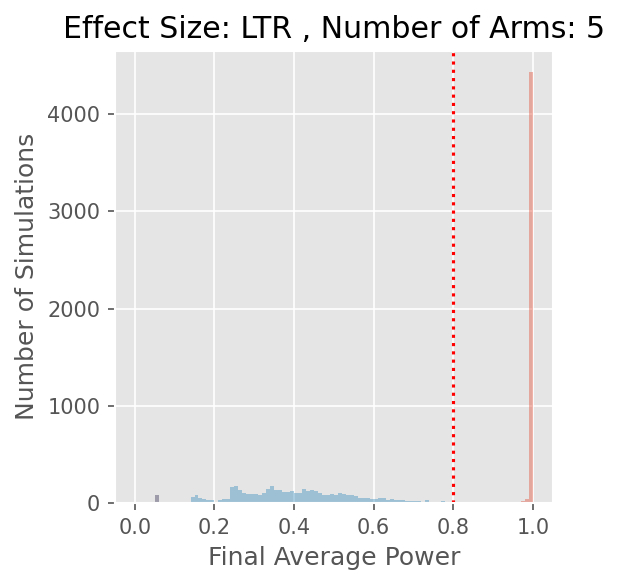} }}
\caption{Comparison of final power between DTS algorithm (blue) and Uniform assignment (red) for LTR dataset where the dotted line indicates effect detected. }
\end{figure}

\begin{figure}[htp]
\centering
\hskip -2.1ex
\mbox{\subfigure{\includegraphics[width=0.32\textwidth]{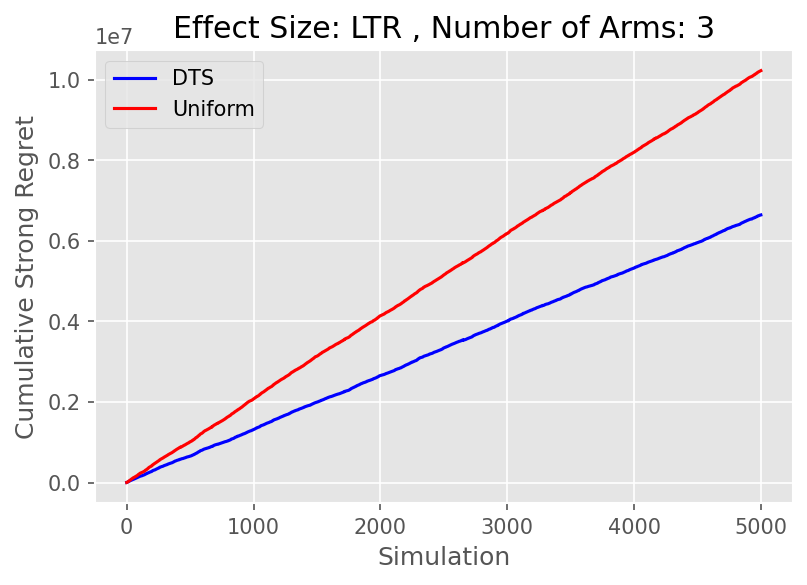}} 
\hskip -0.2ex
\subfigure{\includegraphics[width=0.31\textwidth]{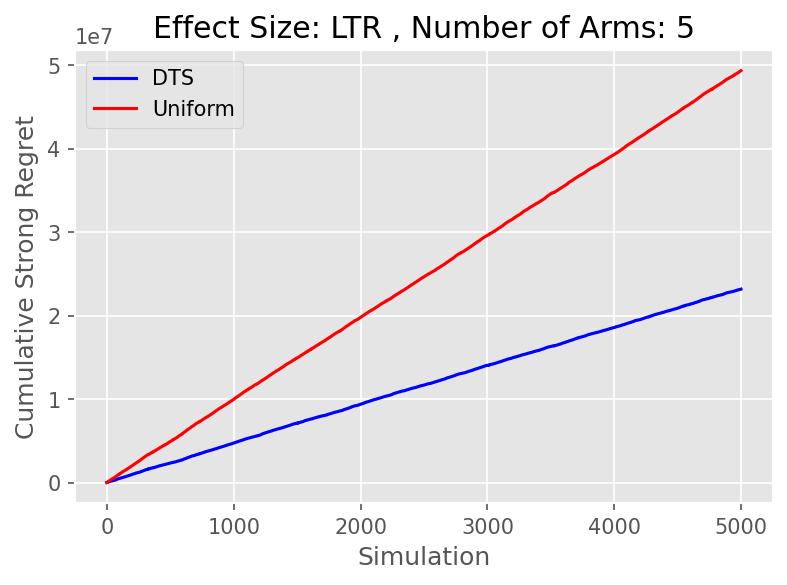} }}
\caption{Comparison of cumulative strong regret over time between DTS algorithm (blue) and Uniform assignment (red) for LTR dataset}
\end{figure}

Finally, we also observed that DTS does a much better job at minimising regret than uniform sampling on the LTR data set as seen in Figure 8.  Overall, we note that both uniform sampling and DTS are useful algorithms in educational contexts depending on the available resources and potential intervention goals. Uniform sampling is more appropriate in cases where finding significant effects is of higher priority and the number of students is small, whereas DTS is more advantageous when better student experience is more important through optimal allocation.
\section{Conclusion}

\label{conclusion}
Adaptive experimentation through bandits presents an important opportunity for researchers, as it allows more participants to receive the probable best intervention, compared to uniform sampling. Although the benefits and drawbacks of experimentation with the Thompson sampling algorithm have been explored, in some cases a dueling bandit paradigm may be more appropriate. To explore the statistical quality of data during randomised experimentation under this framework we conducted an analysis of statistical power, regret, and false positive rates comparing traditional uniform sampling to the DTS algorithm. Our findings indicate that the DTS algorithm shows excellent performance with respect to cumulative regret minimisation but in the case when there is little to no difference between arms, DTS leads to higher Type-I error rates and reduced power. This makes it increasingly difficult to collect data that enables statistical hypothesis tests between conditions, while also balancing this with reward maximization and minimizing the false positive rate. These findings pave the way for researchers to better leverage machine learning for adaptive experimentation, while enabling sound statistical inference. They also present a new opportunity for future work to explore ways of incorporating the DTS algorithm  in adaptive experiments where both reward optimisation and statistical inference are important, possibly through algorithm-driven dynamic interpolation between DTS and uniform exploration for better performance.




\medskip

\small
{
\bibliography{refs.bib}
}

\newpage

\section*{Appendix}

\section*{A. Double Thompson Sampling}


\includegraphics[width=0.78\textwidth]{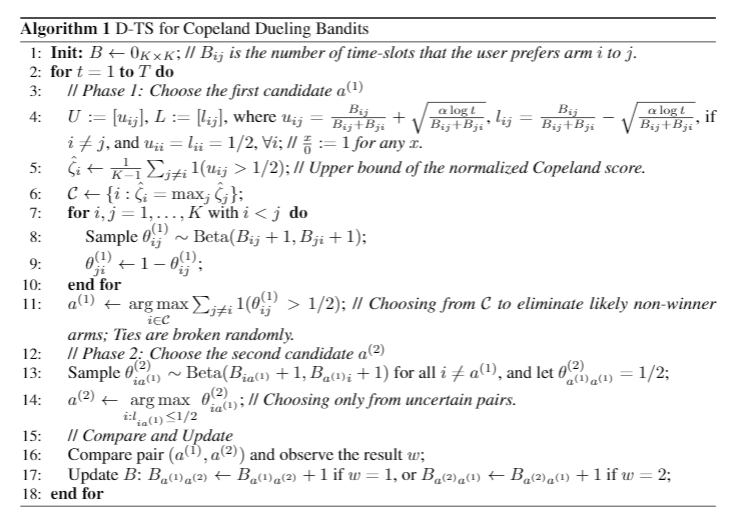}

\end{document}